%% file: main.tex
\pdfoutput=1

\documentclass[11pt]{article}

\usepackage[]{emnlp2024}
\usepackage{times}
\usepackage{latexsym}
\usepackage{csquotes}
\usepackage{tabularx}
\usepackage[normalem]{ulem}
\usepackage{varwidth} 
\usepackage{multirow}
\usepackage{array} 
\usepackage[T1]{fontenc}
\usepackage{pifont}
\usepackage[T5]{fontenc}
\usepackage{tikz}
 
\usepackage[utf8]{inputenc}
\usepackage{soul}
\usepackage{microtype}
\usepackage{graphicx}

\usepackage{arabtex}
\usepackage{utf8}
\newcommand{\ourdataset}{\textit{Casablanca}}
\usepackage{textcomp}  
\usepackage{scalerel}  

\setcode{utf8}

\input{commands}

\title{\ourdataset: Data and Models for Multidialectal Arabic Speech Recognition}

\author{\small 
Bashar Talafha$^{1}$
\hspace{0.02cm}\thanks{\hspace{0.2cm}Corresponding Authors:
\href{btalafha@mail.ubc.ca}{btalafha@mail.ubc.ca}, \href{muhammad.mageed@ubc.ca}{muhammad.mageed@ubc.ca}
}
\\\hspace{0.9cm}\textbf{\small Ahmed O. El-Shangiti$^{2}$}
\\\textbf{\small  Aisha Alraeesi$^{2}$}
\\\hspace{1.3cm}\textbf{\small  El Moatez Billah Nagoudi$^{1}$}
\\\hspace{1.1cm}\textbf{\small  Yasir Ech-chammakhy$^{5}$}
\\\hspace{0.5\textwidth}\textbf{\small Ismail Berrada$^{10}$}
\And
\small  Karima Kadaoui$^{2}$
\\\textbf{\small  Hiba Zayed$^{3}$}
\\\textbf{\small  Hoor Mohamed$^{2}$}
\\\hspace{1.3cm}\textbf{\small  Saadia Benelhadj$^{7}$}
\\\hspace{1.3cm}\textbf{\small  Amal Makouar$^{10}$}
\\\hspace{0.6\textwidth}\textbf{ \small Muhammad Abdul-Mageed$^{1,2,13}$ $^{\ast}$}
\\
\hspace{0.4\textwidth}
\small $^{1}$University of British Columbia,~~ $^{2}$MBZUAI,~~ $^{3}$Birzeit University, ~~$^{4}$JUST,~~ $^{5}$INSEA, $^{6}$Université de Nouakchott, ~~ $^{7}$ESI,
\\
\hspace{0.4\textwidth}
\small $^{8}$Ain Shams Univ., ~~$^{9}$Technische Hochschule Mittelhessen, ~~$^{10}$UM6P, ~~ $^{11}$TEK-UP, ~~$^{12}$Cairo University, ~~$^{13}$Invertible AI
\And
\small  Samar M. Magdy$^{2}$
\\\textbf{\small  Mohamedou Cheikh Tourad$^{6}$}
\\\textbf{\small  Fakhraddin Alwajih$^{1}$}
\\\hspace{0.5cm}\textbf{\small  Hamzah A. Alsayadi$^{8}$}
\\\hspace{1cm}\textbf{\small  Yousra Berrachedi$^{1}$}
\And
\small Mariem Habiboullah$^{9}$
\\\textbf{\small  Rahaf Alhamouri$^{4}$}
\\\textbf{\small  Abdelrahman Mohamed$^{2}$}
\\\hspace{0.5cm}\textbf{\small Walid Al-Dhabyani$^{12}$}
\\\textbf{ \small Mustafa Jarrar$^{3}$}
\And
\hspace{-0.3cm}\textbf{\small Chafei Mohamed$^{11}$}
\\\textbf{\small  Rwaa Assi$^{3}$} 
\\\hspace{-0.4cm}\textbf{\small  Abdellah El Mekki$^{2}$} 
\\\textbf{\small  Sara Shatnawi$^{2}$}
\\\hspace{-0.4cm}\textbf{ \small Shady Shehata$^{2,13}$}
}

\begin{document}

\maketitle
\begin{abstract}
In spite of the recent progress in speech processing, the majority of world languages and dialects remain uncovered. This situation only furthers an already wide technological divide, thereby hindering technological and socioeconomic inclusion. This challenge is largely due to the absence of datasets that can empower diverse speech systems. In this paper, we seek to mitigate this obstacle for a number of Arabic dialects by presenting~\ourdataset, a large-scale community-driven effort to collect and transcribe a multi-dialectal Arabic dataset. The dataset covers eight dialects: Algerian, Egyptian, Emirati, Jordanian, Mauritanian, Moroccan, Palestinian, and Yemeni, and includes annotations for transcription, gender, dialect, and code-switching. We also develop a number of strong baselines exploiting~\ourdataset. The project page for~\ourdataset~is accessible at: \casablancalink.
\end{list}
\end{abstract}

\input{Sections/Introduction}
\input{Sections/Related_work}

\input{Sections/Corpus_Collection/Corpus_Collection}

\input{Sections/Data_Annotation/Data_Annotation}
\input{Sections/Dialects_Description/Dialects_Description}
\input{Sections/Corpus_Statistics/Corpus_Statistics}
\input{Sections/Baseline_models/Baseline_models}

\input{Sections/Conclusion}
\input{Sections/Limitations}
\input{Sections/Ethical_Statement}
\input{ack}

\bibliography{custom}
\bibliographystyle{acl_natbib}

\appendix

\section{Appendix}
\label{sec:appendix}

\input{Sections/APPENDIXES/Appendix}

\end{document}

%% file: commands.tex
\newcommand{\xmark}{\ding{55}}%

\newcommand{\casablancalink}{
\href{https://www.dlnlp.ai/speech/casablanca/}{https://www.dlnlp.ai/speech/casablanca}}

%% file: Sections/Introduction.tex
\section{Introduction}
Self-supervised learning (SSL) has significantly advanced the field of speech processing, impacting everything from speech recognition to speech synthesis and speaker verification. However, the success of these methods heavily relies on the availability of large datasets, which are primarily available for a select few languages. This bias towards resource-rich languages leaves behind the majority of the world's languages \cite{bartelds2023making, talafha2023n, meelen2024end, tonja2024nlp}.
In this work, we report our efforts to alleviate this challenge for Arabic---a collection of languages and dialects spoken by more than 450 million people. We detail a year-long community effort to collect and annotate a novel dataset for eight Arabic dialects spanning both Africa and Asia. This new dataset, dubbed~\ourdataset, is rich with various layers of annotation. In addition to speech transcriptions, we include speaker gender, dialect, and code-switching information. Notably, to the best of our knowledge, some of the dialects included in~\ourdataset~have not been featured in any prior speech or broader NLP research. In addition to describing our dataset, we develop baseline systems for automatic speech recognition (ASR). To summarize, our contributions are as follows: 


\begin{enumerate}
    \item We introduce \ourdataset, the largest fully supervised speech dataset for Arabic dialects, labeled with transcriptions, code-switching, dialect, and gender. 

    
    \item We evaluate SoTA multilingual ASR models and four Arabic-centered Whisper models across the eight dialects in~\ourdataset~to assess their adaptability and performance, particularly in handling the linguistic nuances of Arabic dialectal variation. 

   \item We assess the performance of the best-performing model in code-switching scenarios, analyzing the segments using both the original Latin characters and their transliterated counterparts.

\end{enumerate}

%% file: Sections/Related_work.tex
\section{Related Work}
\label{sec:related-work}

\textbf{Arabic.} Arabic encompasses a diverse array of linguistic varieties, many of which are nearly mutually unintelligible~\cite{watson2007phonology,abdul2024nadi}. This diversity includes three primary categories: Classical Arabic, historically used in literature and still employed in religious contexts; Modern Standard Arabic (MSA), used in media, education, and governmental settings; and numerous colloquial dialects, which are the main forms of daily communication across the Arab world and often involve code-switching~\cite{mageed2020toward,mubarak2021qasr}. The significant differences between these varieties pose challenges in adapting technologies from one variety to another (e.g. MSA to the Yemeni dialect)~\cite{habash2022arabic, talafha2023n}.


\begin{table*}
\centering

\resizebox{\textwidth}{!}{%
\begin{tabular}{lccccc}
\hline
\textbf{}               & \textbf{MGB-2} & \textbf{MGB-3} & \textbf{MGB-5} & \textbf{QASR} & \textbf{\ourdataset}        \\ \hline\\[-0.4cm]
\textbf{Hours}          & 1,200          & 16             & 14             & 2,000         & 48                        \\
\textbf{Dialects} &
  \begin{tabular}[c]{@{}c@{}}(MSA: 78\%+) \\ GLF, LEV, NOR, EGY\end{tabular} &
  EGY &
  MOR &
  \begin{tabular}[c]{@{}c@{}}(MSA: majority) \\ GLF, LEV, NOR, EGY\end{tabular} &
  \begin{tabular}[c]{@{}c@{}}ALG, EGY, JOR, MOR,\\ UAE, PAL, MAU, YEM\end{tabular} \\
\textbf{Dialect Label}  & \xmark              & N/A            & N/A            & \xmark             & 8 labels                  \\[0.2cm]
\textbf{Segmentation}   & lightly        & test: fully    & test: fully    & lightly       & fully                     \\[0.2cm]
\textbf{Transcription}  & lightly        & fully          & fully          & lightly       & fully                     \\[0.2cm]
\textbf{Code-switching} & \xmark              & \xmark              & \xmark              & EN+FR         & EN+FR  (+transliteration) \\[0.2cm]
\textbf{Gender}         & \xmark              & \xmark              & \xmark              & $\approx$82\% data     & 100\% data                \\[0.2cm]
\hline
\end{tabular}
}
\caption{~\ourdataset~in comparison to notable Arabic speech datasets. \textbf{Lightly}: lightly supervised (labeling is performed using a pre-trained model). \textbf{Fully}: fully supervised (all annotations are carried out manually by humans). \textbf{Test: fully}: {only the test set is labeled manually} .\xmark: does not support. \textbf{N/A}: not applicable as those datasets have one dialect only. \textbf{EN}: English. \textbf{FR}: French. \textbf{+transliteration}: code-switching words are written in both Latin and Arabic scripts.}
\label{tab:datasets}
\end{table*}

\textbf{Arabic ASR data.}
Early efforts to develop Egyptian Arabic speech datasets began in 1996 with the \textit{CallHome} task~\cite{pallett2003look} under the National Institute of Standards and Technology's (NIST) evaluations, focusing on the Egyptian and Levantine dialects. In 2006, the DARPA-led Global Autonomous Language Exploitation (GALE) \cite{galeprog} and the Spoken-Language Communication and Translation System for Tactical Use (TRANSTAC) programs \cite{transtac} aimed to develop Iraqi dialect dataset, driven by U.S. military needs~\cite{olive2011handbook}. 
The Multi-Genre Broadcast (MGB) Challenge \st{has} later introduced several datasets aimed at advancing speech recognition, speaker diarization, alignment, and dialect identification using content from TV and YouTube. 
\textit{MGB-2}~\cite{ali2016mgb} provides 1,200 hours of speech with lightly supervised transcriptions, derived from Aljazeera Arabic news broadcasts with MSA making up 78\%\footnote{The updated version of MGB-2 reported 78\%, while the old one reported 70\%~\cite{mubarak2021qasr}.} of the total content. 
\textit{MGB-3}~\cite{ali2017speech} compiles video clips from Egyptian YouTube channels while
MGB-5~\cite{ali2019mgb} focuses on Moroccan Arabic ASR. 
Additionally, the \textit{QASR} project~\cite{mubarak2021qasr}, sourced from Aljazeera’s archives between 2004 and 2015, features over 4,000 episodes across various topics, including extensive code-switched transcriptions from multiple dialects. Further details of the MGB and QASR datasets are provided in Table~\ref{tab:datasets}.

\textbf{Non-Arabic ASR data.} Similar efforts exist for collecting diverse speech datasets across various language varieties and dialects. For instance, STT4SG-350~\cite{pluss2023stt4sg} introduces a Swiss German corpus divided into seven dialect regions, annotated with Standard German transcriptions. AfriSpeech~\cite{olatunji2023afrispeech} also offers 200 hours of Pan-African English speech, featuring 67,577 audio clips from speakers across 13 countries, encompassing 120 indigenous accents for both clinical and general ASR applications. The ManDi Corpus~\cite{zhao2022mandi} provides a detailed spoken database of regional Mandarin dialects and Standard Mandarin, with 357 recordings totaling about 9.6 hours from 36 speakers across six major regions.
Additional information on Arabic ASR can be found in Appendix \ref{app:ar-speech}.

 \textbf{\ourdataset~in comparison.}
\ourdataset~is the largest fully supervised Arabic dialects dataset with 48 hours of human-transcribed data, surpassing MGB-3 and MGB-5. Although MGB-2 and QASR are larger in size, they utilize light supervision (using  ASR systems for transcribing and aligning human transcripts) rather than manual transcriptions. This light supervision method accounts for potential inaccuracies in human transcripts, such as omissions, errors, and variations from factors like corrections, spelling errors, foreign language use, and overlapping speech, leading to possible mismatches between the transcriptions and actual spoken content~\cite{mubarak2021qasr}.
~\ourdataset~ is also the most fine-grained and diverse corpus available: while datasets such as MGB-2 and QASR focus on broad regional dialects like the Gulf, the Levant, and North Africa (including Egypt), ~\ourdataset~ targets country-level variation focusing on eight countries belonging to different areas in the Arab world. 
To the best of our knowledge, our dataset is also the first to introduce zero-resourced dialects in addition to the low-resource ones (specifically the Emirati, Yemeni, and Mauritanian dialects), thus filling a significant need in the research landscape. 
%
Furthermore,~\ourdataset~is rich with several layers of annotation: beyond \textit{speech transcription},  each segment is also labeled with speaker \textit{gender} and \textit{country}, which provide valuable demographic information and can be exploited for downstream tasks involving gender and dialect identification. 
Table~\ref{tab:datasets} provides a comparison between~\ourdataset~and a number of notable Arabic datasets. 
Finally, with \ourdataset, 
we are advancing the benchmarking efforts to encompass eight dialects and include evaluations on four multilingual models: Whisper~\cite{radford2023robust} (both versions 2 and 3), SeamlessM4T~\cite{barrault2023seamlessm4t}, and MMS~\cite{pratap2023scaling} under zero-shot and Arabic-enhanced\footnote{Further finetuned on Arabic data.} settings. This expansion strengthens our analysis by incorporating advanced models, offering a comprehensive evaluation of their capacity to handle diverse dialects.



%% file: Sections/Corpus_Collection/Corpus_Collection.tex
\section{Corpus Collection}
\label{sec-corpus-collection}
\input{Sections/Corpus_Collection/Data_Selection}

\input{Sections/Corpus_Collection/Data_Segmentation}


%% file: Sections/Corpus_Collection/Data_Selection.tex
\subsection{Data Selection}
\begin{figure*}[t]
  \begin{center}
    \includegraphics[width=\linewidth]{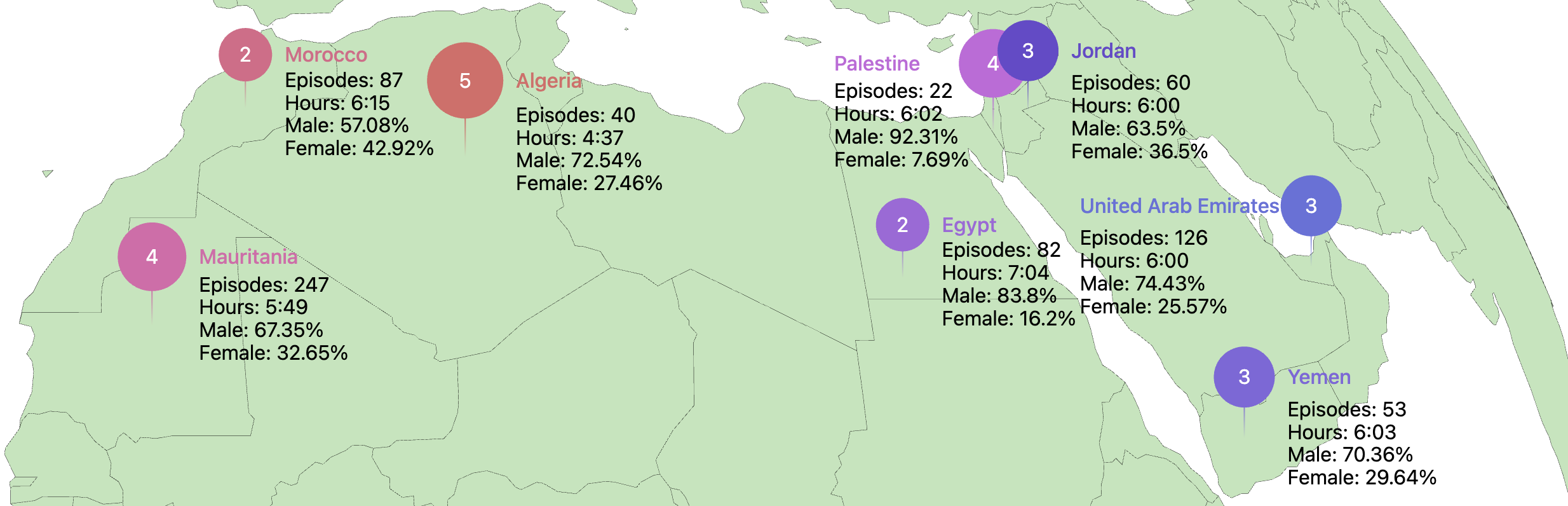}
  \end{center}
\caption{Geographic distribution of participants and data in~\ourdataset. \textbf{Pins} on each country represent the number of participants per dialect. \textbf{Episodes} denotes the number of selected episodes. \textbf{Hours} refer to the total hours of transcription per dialect.  \textbf{Male} and \textbf{Female} are percentages of male and female speaker coverage over dialects. } 
\label{fig:map}
\end{figure*}
We assembled a team of 15 native speakers (each with a research background) and assigned them the task of manually curating a list of YouTube episodes from TV series that represent the dialects of their countries.
To ensure diversity, we instruct them to include a variety of actors and geographical settings\footnote{This involves diverse genders, ages, speaking styles, and locations reflecting various sub-dialects within the country.}.
We manually verified that each episode is over 15 minutes in length and removed introductory videos, such as trailers, to eliminate redundancy. 
Due to copyright restrictions on the original YouTube videos, we follow the approach by \newcite{uthus2024youtube, ali2019mgb, ali2017speech} and do not provide them directly. Instead, we make available the YouTube URLs, timestamps, and annotations.
The copyright remains with the original video owners and data we release will be exclusively for research purposes.\footnote{The project page for{~\ourdataset} is accessible at: 
\casablancalink.} 

%% file: Sections/Corpus_Collection/Data_Segmentation.tex
\subsection{Data Segmentation}
We segment the episodes into shorter utterances, thereby simplifying transcription and enabling task distribution among annotators for a more streamlined process. 
We use the voice activity detection model (VAD) of~\citet{Bredin2021, Bredin2020}, available through the \textit{pyannotate}, to detect speech and remove non-speech segments such as music\footnote{We utilize the model with its default hyperparameters ($onset$: $0.8104$, $\mathit{offset}$: $0.4806$,
                $min\_duration\_on$: $0.055$, $min\_duration\_\mathit{off}$: $0.097$).}. 
We then use \textit{AudioSegment}\footnote{\href{https://github.com/jiaaro/pydub}{https://github.com/jiaaro/pydub}} to extract the identified speech segments.
We refer to these extracted audio segments as `\textbf{snippets}'. It is important to note that an output snippet may contain multiple utterances, often involving various speakers. We put the snippets on the LabelStudio platform \cite{tkachenko2020label} for annotation. See more details about annotation in Appendix~\ref{sec-annotation-tool}.






%% file: Sections/Data_Annotation/Data_Annotation.tex
\section{Data Annotation}
\label{sec:data-annotation}

\input{Sections/Data_Annotation/Participants}
\input{Sections/Data_Annotation/Tasks}

%% file: Sections/Data_Annotation/Participants.tex
\subsection{Annotators}




Our community-driven dataset, \ourdataset, is created with the help of 27 annotators from the Arab world, each annotating their respective dialects.
All annotators either have or are pursuing graduate degrees in natural language processing, making them well-positioned for the task. We involve at least two annotators per dialect, each coming from a different region within the respective country for an enhanced knowledge of sub-dialects\footnote{In the literature, these sub-dialects are sometimes referred to as ``micro-dialects"~\cite{mageed2020toward}.}, which adds a layer of linguistic richness and diversity to the orthographic representation of each dialect. Table~\ref{tab:variations} (Appendix \ref{Inter-dialect}) illustrates lexical variation within the eight dialects in~\ourdataset, showcasing its linguistic diversity.

%% file: Sections/Data_Annotation/Tasks.tex
\subsection{Tasks}
\label{subsec-tasks}
We provided annotators with written guidelines explaining the annotation tasks. During weekly meetings with team members, we discussed, improved, and iteratively extended these guidelines. Annotators are also able to communicate with one another and ask questions through a Slack channel dedicated to the project. The main annotation tasks are.

\noindent





\textbf{\textit{Task 1: Segment Selection}}
We introduced three annotation options as shown in Figure \ref{fig:msadiaothers}: \textit{Dialect} for dialect-specific content, \textit{MSA} for Modern Standard Arabic, and \textit{Other} for segments containing non-verbal sounds. Selected segments, whether dialectal or MSA, are required to be "clear segments".
They must feature only one speaker to avoid voice overlap, be audibly clear and transcribable despite potential background noise, and contain a minimum of three words without surpassing 30 seconds in length. Moreover, each segment must capture the complete utterance, from beginning to end, accurately representing every phoneme component of the first and last words to preserve speech boundaries. 
%
%
%
%
%
%
%
%
%

\textit{\textbf{Task 2: Transcription}}
Given the absence of a standardized orthographic system for Arabic dialects, we asked annotators to transcribe in the manner they usually write in their daily lives. 
Furthermore, for a faithful representation of the speech signal, we encouraged the incorporation of Tanweens and Hamzat\footnote{Tanween refers to the doubling of a vowel at the end of a word, indicated by diacritic marks, enhancing the noun's indefinite status in Arabic. Hamza represents a glottal stop, marked by its diacritic, crucial for words disambiguation \cite{el2004phonetization}.} in the transcriptions.
We also asked annotators to render numbers in alphabetical format (e.g., \<انا عاوز عشرين بطاقة>) instead of numerical symbols (e.g., \<انا عاوز 20 بطاقة>), since this allows for reflecting inflections these numbers can have (e.g., \<عشرين> vs. \<عشرون>).
For code-switching (CS), we asked annotators to provide two versions of the transcript, one with the foreign words in Arabic script (e.g., \<بروفيشينال>) and another in Latin script (e.g., "professional"); see Table \ref{tab:cs} in Appendix \ref{app-cs}.

\textit{\textbf{Task 3: Gender}}
Annotators label speaker gender based on perceived biological sex\footnote{This acknowledges differences between biological sex and gender identity.} from the set \{\textit{male, female}\}. This makes our dataset suited for studying gender-specific speech patterns across dialects.



\textit{\textbf{Task 4: Validation}}
In this task, each team engages in a peer validation process, with annotators reviewing and ensuring the accuracy of one another's transcriptions, focusing on correcting spelling errors while preserving dialectal orthographic variations.
Our annotation process utilized an agile methodology~\cite{cohen2004introduction} with work divided into weekly sprints, allowing for focused objectives and regular review sessions to refine strategies. 
We also gave annotators a guideline document\footnote{Our annotation guidelines are available at the project page: \casablancalink.} and a document on special cases to standardize dialect scenarios and document linguistic variations. See Appendix \ref{apx:special-cases} for examples. Overall, the annotation project ran for a total duration of six months.

%% file: Sections/Dialects_Description/Dialects_Description.tex
\section{Dialects Description}
\label{sec:dialect-desc}

\ourdataset~is a detailed collection of around 48 hours of data covering eight Arabic dialects from regions like the Levant, Gulf, Yemen, and North Africa, including Algerian, Egyptian, Emirati, Jordanian, Mauritanian (Hassaniya), Moroccan, Palestinian, and Yemeni.
\ourdataset~involves sub-dialects from these countries as well. In addition, to the best of our knowledge, we are among the first to offer annotated data for the less-represented Emirati, Mauritanian, and Yemeni dialects, addressing a gap in linguistic research.


%% file: Sections/Corpus_Statistics/Corpus_Statistics.tex
\section{Corpus statistics}
\label{sec-corpus-statistics}
\textbf{Episode Coverage.}
As spelled out earlier, we annotate approximately 48 hours of content across eight dialects. The average annotation duration per episode is about four minutes, constituting roughly 14.71\% of the average episode length. Dialects represented by a larger number of episodes typically exhibit lower per-episode annotation durations.
This distribution allows annotators to engage with a more diverse range of content. For instance, Mauritanian episodes, totaling 247, feature an average of only one minute and 25 seconds ($8.23\%$) of annotation per episode. Conversely, the Palestinian subset, with 22 episodes, averages 16 minutes and 30 seconds per episode, which is about $53.72\%$ of the total episode length.\footnote{Despite our efforts, we could not acquire more episodes where the Palestinian dialect is not mixed with other dialects.}






\begin{table*}[t!]
\resizebox{\textwidth}{!}{%

\begin{tabular}{lccccclllcc}
\hline 
\textbf{Dialect} &
\textbf{Total Dur} &
  \textbf{Avg Dur} &
  \textbf{AVT } &
  \textbf{U-Wds} &  
  \textbf{Avg U-Wds/hr} &
  \textbf{Snippets} &
  \textbf{Segments} &
  \textbf{Skips} &
  \textbf{Avg WPS / CPS} &
  \textbf{CS} 
  \\
  \hline 
\textbf{Algeria} &  4:37:35  & 4.15 & 8.41 & 11,085 & 3,518 &2,537 & 4,013 & 769 & 2.662 / 10.723 & 586\\
\textbf{Egypt} &  7:04:16   & 4.29 & 10.67 & 16,080 & 3,981 &2,962 & 5,937 & 715 & 2.858 / 13.165 & 72\\
\textbf{Jordan}  &  6:00:16  & 4.23 & 5.71 & 13,145 & 3,653&4,255 & 5,105 & 5,257 & 1.286 / 6.142 & 52\\
\textbf{Mauritania} &  5:49:40 & 3.67 & 5.83 & 12,835 & 3,605 &3,099 & 5,325 & 5,556 & 1.631 / 7.170 & 36\\
\textbf{Morocco} &  6:15:02  & 3.54 & 10.83 & 15,469 & 4,458&4,119 & 6,358 & 504 & 3.206 / 15.728 & 598\\
\textbf{Palestine} & 6:02:59 & 5.30 & 11.30 & 13,405 & 3,628&2,543 & 4,107 & 720 & 2.264 / 10.612 & 50\\
\textbf{UAE}   &  6:00:06   & 4.25 & 9.57 & 13,067 & 3,565&2,780 & 5,087 & 853 & 2.362 / 10.954 & 59\\
\textbf{Yemen}  & 6:03:26   & 4.49 & 6.85 & 16,140 & 4,175&2,991 & 4,861 & 3,825 & 1.517 / 7.393 & 1\\
\hline
\textit{\textbf{Total}} &  \textbf{47:53:20    }& 4.24 & 8.64 & 85,176 & 3,822.9 & 25,286 & 40,793 & 18,199 & 2.223 / 10.235 &  1,454\\
\hline
\end{tabular}
}


\caption{Distribution of data in ~\ourdataset. \textbf{Total Dur}: total duration for each dialect. \textbf{Avg Dur}: total duration divided by number of segments. \textbf{AVT}: average transcript length. \textbf{U-Wds}: number of unique words. \textbf{Avg U-Wds/hr}: average number of unique words per hour. \textbf{Skips}: number of skipped snippets. \textbf{WPS}: words per second. \textbf{CPS}: characters per second. \textbf{CS}: Number of code-switching segments. For \textit{\textbf{Total}}, we take the average for average columns and sums for other columns.
}
\label{tab:stat}
\end{table*}

\textbf{Average Duration.} As detailed in Table \ref{tab:stat}, the average duration of segments across all dialects stands at $4.24$ seconds, with the Moroccan having the shortest average duration and the Palestinian the longest. We define the speed rate as the average number of words per second (WPS) and the average number of characters per second (CPS). Interestingly, based on our analysis of the episodes, the Moroccan dialect stands out as the fastest spoken dialect in~\ourdataset, both in terms of WPS and CPS with $3.2$ WPS and $15.7$ CPS, respectively. Conversely, Jordanian dialect is the slowest in our dataset, yielding $1.2$ WPS and $6.14$ CPS\footnote{Fastest to slowest: Morocco > Egypt >  Algeria >  UAE >  Palestine >  Mauritania >  Yemen > Jordan. Although these observations are useful, we acknowledge they may be particular to our own dataset and hence should not be generalized.}.
 
The average transcript length across all dialects is $8.64$ words, with Jordanian transcripts being the shortest and Palestinian the longest. These differences, even between closely related dialects, stem from episode script lengths and annotator preferences for word separation, including prefixes and suffixes.
For instance, in the Jordanian dialect, the phrase ("\textit{I sent it to her}") transcribed by some annotators as a single word: ("\<بعثتلهاياها>"), while others split it into two: ("\<بعثتلها اياها>") or even three words: ("\<بعثت الها اياها>"). 
This highlights the subjectivity among annotators across the various dialects that influence word count and segment length differences. 
This subjectivity, in addition to the episodes' topic diversity, influence the unique word count per dialect as detailed in Table~\ref{tab:stat}. For all dialects combined, the unique word count is 85,176 words.
On a country level, the Morrocan dialect has the highest number of unique words per hour with 4,458 words, while the Algerian dialect has the smallest at 3,518 words.
This indicates that, besides Moroccan being the fastest dialect, it also has the greatest word diversity compared to other dialects.
\textbf{Code-Switching.} Among all dialects in~\ourdataset, Algerian and Moroccan demonstrate a notably high usage of code-switching. Namely, as Table~\ref{tab:stat} shows, these dialects feature $500$+ segments with code-switching. 
These North African dialects, in addition to Mauritanian, uniquely blend French into their code-switching. Other dialects in our dataset, such as Egyptian and Jordanian, involve switching into English. This linguistic diversity mirrors the historical colonial impact on languages in these regions. Overall, \ourdataset~includes 234 English code-switching segments (totaling $\approx22$ minutes) and 1,220 French code-switching segments (one hour and 44 minutes). Examples are shown in Table~\ref{tab-csstat} in Appendix \ref{app-cs}.
Conversely, we observe less code-switching in the other dialects. We suspected this may be due to episodes from other countries being relatively older as use of code-switching has become more prevalent among younger Arab generations~\cite{brown2005encyclopedia}. To test this hypothesis, we manually labeled the episodes for their time coverage. We found the following: Egypt (1997-2018), Jordan (1985-2000), and UAE (1995-2009) with 72, 52, and 59 code-switching instances, respectively. In contrast, newer episodes show higher instances: Algeria (2004-2017), and Morocco (2016-2018) with 586 and 598 cases, respectively. To summarize, our analysis shows that (i) French code-switching is more common than English and, even within the same dialect, (ii) newer episodes involve more code-switching than older ones.



\textbf{Gender Bias.} Despite our efforts to balance gender representation, a clear male dominance is observed across all dialects as demonstrated in Figure~\ref{fig:map}. The disparity is most notable in the Palestinian dialect, where male voices constitute 92.31\%, leaving a mere 7.69\% for female representation. 
In contrast, the Moroccan dialect exhibits a more gender balanced setup (with 57.08\% male and 42.92\% female). We now describe baseline models we developed exploiting our dataset.

%% file: Sections/Baseline_models/Baseline_models.tex
\section{Baseline models}
\label{sec:baseline-models}
\input{Sections/Baseline_models/ASR}

%% file: Sections/Baseline_models/ASR.tex

\begin{table*}[t!]
\resizebox{\textwidth}{!}{%
\begin{tabular}{lcccccccc}
\hline
\textbf{} &
  \multicolumn{2}{c}{\textbf{whisper-lg-v2}} &
  \multicolumn{2}{c}{\textbf{whisper-lg-v3}} &
  \multicolumn{2}{c}{\textbf{seamless-m4t-v2-large}} &
  \multicolumn{2}{c}{\textbf{mms-1b-all}} \\ \hline
\multicolumn{1}{c}{} &
  \multicolumn{1}{c}{\textbf{- pre-proc}} &
  \multicolumn{1}{c}{\textbf{+ pre-proc}} &
  \multicolumn{1}{c}{\textbf{- pre-proc}} &
  \multicolumn{1}{c}{\textbf{+ pre-proc}} &
  \multicolumn{1}{c}{\textbf{- pre-proc}} &
  \multicolumn{1}{c}{\textbf{+ pre-proc}} &
  \multicolumn{1}{c}{\textbf{- pre-proc}} &
  \multicolumn{1}{c}{\textbf{+ pre-proc}} \\ \hline
  
\textbf{Algeria} &
82.61	/ 38.95	&
\textbf{80.47	/ 36.82} &
83.49	/ 40.47	&
84.14	/ 39.99	&
101.18	/ 58.58	&
94.18	/ 53.56	&
93.01	/ 43.68	&
92.55	/ 42.62 \\
\textbf{Egypt} &
61.99	/ 26.38	&
52.38 /	21.71	&
59.11 /	24.77	&
\textbf{48.95	 / 19.86} &	
61.82 / 	29.83	 &
49.75 / 	24.47	&
88.54	 / 43.59	&
85.84	 / 40.58 \\
\textbf{Jordan} &
49.47 / 	16.34 &
41.13 / 	13.64	&
48.44 / 	16.18	&
39.68	 / 13.47	&
47.94 / 	15.84	&
\textbf{39.24 / 	13.12} &
81.46 / 	33.02 &
78.54 / 	31.03  \\
\textbf{Mauritania} &
87.85 / 	52.34 &
85.74 / 	49.76 &
87.44 / 	50.19 &
\textbf{85.68 / 	48.08} &
91.57 / 	55.41 &
88.39 / 	51.59 &
94.36 / 	50.25 &
93.71 / 	48.99\\
\textbf{Morocco} &
88.55 / 	46.57 &
84.52 / 	44.02 &
87.2 / 	44.41 &
\textbf{83.05 / 	42.09} &
95.18 / 	58.29 &
91.01 / 	54.97 &
96.91 / 	49.01 &
95.45 / 	47.34\\
\textbf{Palestine} &
57.06 / 	20.02 &
\textbf{48.64 / 	17.24} &
58.02 / 	21.05 &
50.2 / 	18.38 &
56.78 / 	20.74 &
48.92 / 	18.13 &
83.14 / 	33.07 &
80.18 / 	30.82\\
\textbf{UAE} &
61.82 / 	22.93 &
\textbf{52.03 / 	19.15} &
62.31 / 	24.04 &
52.88 / 	20.37 &
63.94 / 	26.22 &
54.76 / 	22.71 &
85.4 / 	36.81 &
82.11 / 	34.18\\
\textbf{Yemen} &
71.31 / 	29.8 &
60.65 / 	24.49 &
69.94 / 	28.17 &
\textbf{59.45 / 	23.19} &
73.65 / 	32.55 &
62.72 / 	27.43 &
86.73 / 	38.55 &
81.64 / 	34.36\\\hline
\textbf{\textit{AVG}} &
70.08 / 	31.66 &
63.195 / 	28.35 &
69.49 / 	31.16 &
\textbf{63.00 / 	28.17} &
74.00 / 	37.18 &
66.12 / 	33.24 &
88.69 / 	40.99 &
86.25 / 	38.74\\ \hline
\end{tabular}
}
\caption{Results for dialect evaluation, scenario-1 on the Test set. Results are reported in WER and CER (/ separated). \textbf{pre-proc:} preprocessing (+ with, - without). }
\label{tab-sc-1}
\end{table*}

\begin{table*}[t!]
\resizebox{\textwidth}{!}{%
\begin{tabular}{lcccccccc}
\hline
\textbf{} &
  \multicolumn{2}{c}{\textbf{whisper-msa}} &
  \multicolumn{2}{c}{\textbf{whisper-mixed}} &
  \multicolumn{2}{c}{\textbf{whisper-egyptian}} &
  \multicolumn{2}{c}{\textbf{whisper-moroccan}} \\ \hline
\multicolumn{1}{c}{} &
  \multicolumn{1}{c}{\textbf{- pre-proc}} &
  \multicolumn{1}{c}{\textbf{+ pre-proc}} &
  \multicolumn{1}{c}{\textbf{- pre-proc}} &
  \multicolumn{1}{c}{\textbf{+ pre-proc}} &
  \multicolumn{1}{c}{\textbf{- pre-proc}} &
  \multicolumn{1}{c}{\textbf{+ pre-proc}} &
  \multicolumn{1}{c}{\textbf{- pre-proc}} &
  \multicolumn{1}{c}{\textbf{+ pre-proc}} \\ \hline
\textbf{Algeria} &
  87.86  /  48.31 &
  87.82 / 48.20 &
  129.63 / 79.63 &
  129.77 / 79.68 &
  86.68 / 35.80 &
  86.75 / 35.70 &
  \textbf{74.39} / 29.50 &
  74.40 / \textbf{29.42} \\
\textbf{Egypt} &
  67.68 / 35.22 &
  67.56 / 35.22 &
  97.31 / 63.87 &
  97.24 / 63.79 &
  49.58 / 19.33 &
  \textbf{49.49 / 19.24} &
  74.82 / 34.83 &
  74.78 / 34.80 \\
\textbf{Jordan} &
  61.18 / 23.43 &
  51.93 / 20.43 &
  78.15 / 40.34 &
  68.89 / 37.84 &
  56.11 / 18.15 &
  \textbf{46.45 / 15.02} &
  72.79 / 27.12 &
  64.87 / 24.32 \\
\textbf{Mauritania} &
  88.02 / 47.5 &
  88.02 / 47.44 &
  114.39 / 78.02 &
  114.43 / 78.09 &
  \textbf{87.08 / 43.32} &
  87.11 / 43.35 &
  89.93 / 45.16 &
  89.93 / 45.17 \\
\textbf{Morocco} &
  88.06 / 46.37 &
  88.03 / 46.37 &
  120.59 / 77.44 &
  120.61 / 77.45 &
  84.85 / 37.22 &
  84.85 / 37.20 &
  61.58 / 21.25 &
  \textbf{61.57 / 21.24} \\
\textbf{Palestine} &
  68.06 / 28.90 &
  59.78 / 26.00 &
  76.92 / 36.81 &
  67.90 / 34.25 &
  63.70 / 22.31 &
  \textbf{54.13 / 19.13} &
  76.83 / 30.15 &
  69.42 / 27.36 \\
\textbf{UAE} &
  74.24 / 35.37 &
  64.54 / 31.79 &
  104.60 / 60.20 &
  96.95 / 57.99 &
  67.45 / 24.48 &
  \textbf{56.58 / 20.27} &
  78.37 / 31.51 &
  70.41 / 27.95 \\
\textbf{Yemen} &
  74.71 / 36.08 &
  69.55 / 33.15 &
  96.01 / 54.81 &
  91.58 / 53.19 &
  70.49 / 28.07 &
  \textbf{64.96 / 24.83} &
  79.13 / 33.89 &
  75.09 / 31.00 \\\hline
\textbf{\textit{AVG}} &
76.225 /	37.6475 &
  72.15 / 36.08 &
  102.20 / 61.39 &
  98.42 / 60.29 &
  70.74 / 28.58 &
  \textbf{66.29 / 26.84} &
  75.98 / 31.68 &
  72.56 / 30.16 \\ \hline
\end{tabular}
}
\caption{Results for dialect evaluation, scenario-2 on the Test set. Results are reported in WER and CER (/ separated). \textbf{pre-proc:} preprocessing (+ with, - without). }
\label{tab-sc-2}
\end{table*}

We split~\ourdataset~into Train, Dev, and Test, keeping the latter two splits each at one hour of the data per country. We perform a number of ASR experiments on the Dev and Test splits of~\ourdataset\footnote{In this work, we do not use the Train splits in any experiments.}. First, we evaluate general speech models under a zero-shot condition. Then, we evaluate models that were finetuned on MSA or other dialects. Finally, we report experiments on our code-switched data only. We report results in WER and CER, both with and without preprocessing of the data. Details of our preprocessing pipeline are in Appendix~\ref{apx:preproc}.

\subsection{Evaluation of General Models}\label{subsec:zse}
We evaluated SoTA multilingual speech models on each dialect to understand their generic adaptability and performance across the eight dialects. Particularly, we evaluated two versions of Whisper~\cite{radford2023robust} (\textit{whisper-large-v2}\footnote{\href{https://huggingface.co/openai/whisper-large-v2}{https://huggingface.co/openai/whisper-large-v2}} and \textit{whisper-large-v3}\footnote{\href{https://huggingface.co/openai/whisper-large-v3}{https://huggingface.co/openai/whisper-large-v3}}, 1550M),  SeamlessM4T~\cite{barrault2023seamlessm4t} (\textit{seamless-m4t-v2-large}\footnote{\href{https://huggingface.co/facebook/seamless-m4t-v2-large}{https://huggingface.co/facebook/seamless-m4t-v2-large}}, 2.3B), and MMS~\cite{pratap2023scaling} (\textit{mms-1b-all}\footnote{\href{https://huggingface.co/facebook/mms-1b-all}{https://huggingface.co/facebook/mms-1b-all}}, 1B)\footnote{We could not evaluate Google USM model~\cite{zhang2023google} since it was not available as of the time of our writing this paper.}. For this scenario, we report WER and CER of four different multilingual models on the eight novel dialects, which we hypothesize may not have been incorporated into the training data of these models. As shown in Table~\ref{tab-sc-1}, all models exhibited high WER and CER across each dialect, indicating their inability to effectively generalize to entirely novel conditions. On average, \textit{whisper-large-v3} recorded lower WER and CER compared to other models, both with preprocessing (63 WER and 28.17 CER) and without (69.49 WER and 31.16 CER). In terms of dialects, without any preprocessing, only on the Jordanian dialect we achieved a WER of less than 50, as recorded by both Whisper models and SeamlessM4T. After preprocessing, the Palestinian and Egyptian dialects approached a WER of around 50 with these models. On average, \textit{mms-1b-all} yielded the lowest performance compared to others, which can be attributed to the significant difference in domains between MMS data, a closed domain focusing on religious texts in MSA, and the Youtube series, an open domain featuring dialectal content.


\subsection{Evaluation of Dedicated Models}\label{subsec:dme} 
\begin{figure}[h!]
  \begin{center}
    \includegraphics[width=\linewidth]{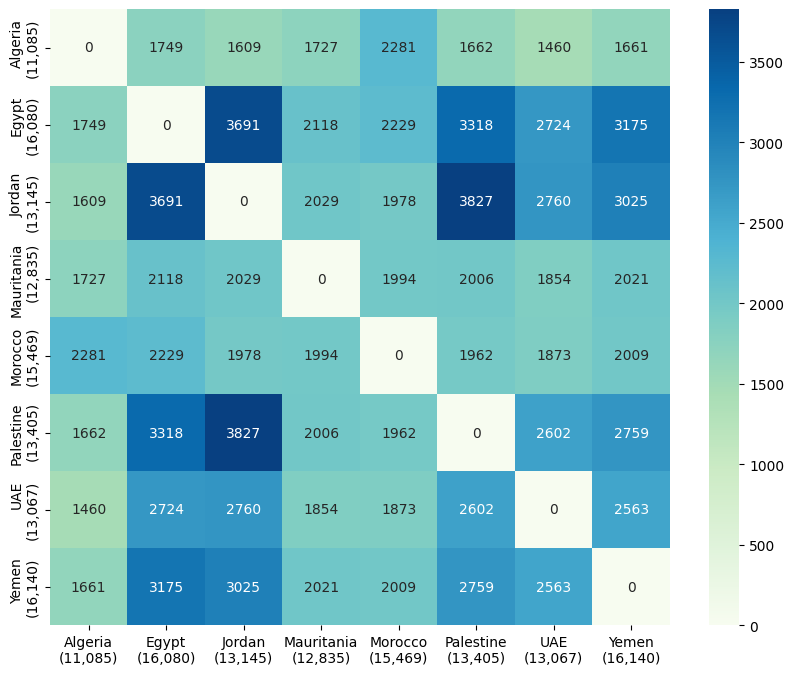}
  \end{center}
\caption{Vocabulary intersection in~\ourdataset. "$0$" denotes no intersection with the dialect itself. Numbers under the country name denote the vocab size.
 }
\label{fig:vocabs}
\end{figure}

Here we evaluate models that were finetuned by~\newcite{talafha2023n} on MSA, Egyptian, and Moroccan. Since the models were not released, we follow the same approach in ~\newcite{talafha2023n} and regenerate\footnote{Regenerate here means that we did the same finetunings in \cite{talafha2023n}}
four Arabic Whisper models
based on whisper-large-v2: 
\textit{whisper-msa} on Common Voice 11.0\footnote{\href{https://huggingface.co/datasets/mozilla-foundation/common_voice_11_0}{https://huggingface.co/datasets/mozilla-foundation/common\_voice\_11\_0}} (CV11) for MSA, \textit{whisper-mixed} on MGB-2 targeting a blend of MSA and dialects, \textit{whisper-egyptian} on MGB-3 focused on the Egyptian dialect, and \textit{whisper-moroccan} on MGB-5 for the Moroccan dialect. 
Then, we evaluate these models on all dialects in~\ourdataset.
As reported in Table~\ref{tab-sc-2}, \textit{whisper-egyptian} is notably superior for all dialects except Moroccan and Algerian.
The superior performance of \textit{whisper-egyptian} can be attributed to its enhanced likelihood of predicting dialectal words, a result of its fine-tuning, compared to \textit{whisper-msa}. Additionally, \textit{whisper-egyptian} is closely aligned with conversational domains that focus on everyday topics, a characteristic shared across all dialectal datasets. In comparison with \textit{whisper-moroccan}, from a vocabulary perspective, as shown in Figure \ref{fig:vocabs}, the Egyptian dialect shares more vocabulary with Yemen, Jordan, UAE, Egypt, Palestine, and Mauritania than with the Moroccan dialect.
Conversely, the Moroccan and Algerian dialects demonstrate a closer vocabulary alignment since these two North African dialects share more linguistic similarities than with other dialects. 
This correlation is consistent with the patterns observed in our experimental results.
Therefore, \textit{whisper-moroccan} performed better for Moroccan and Algerian compared to other models. Despite having the most extensive Arabic content (MGB-2 1200hrs), \textit{whisper-mix} model showed the weakest performance overall. This is attributed to two main reasons: firstly, the data was recorded in studio settings (\textit{Aljazeera.net}); and secondly, the content domain of the MGB-2 dataset (which includes politics, economy, society, culture, media, law, and science) differs significantly from daily conversation topics. This suggests that even though over 70\% of the MGB-2 data is MSA, the remainder in dialects also does not accurately represent everyday speech, leaning more towards these specific close-domains. The evidence from the dialectal models supports the argument, showing that the MGB-3 and MGB-5 datasets, which were collected from YouTube (not including TV series), represent a wider range of real-life domains. Although these datasets are smaller in size compared to MGB-2, the relevance of the domain directly influenced their performance. This effect is also noticeable in the comparison of the whisper-msa and whisper-mixed models. Both performed well with MSA, as reported in~\newcite{talafha2023n}, yet \textit{whisper-msa} yielded better outcomes on dialects than \textit{whisper-mixed}, even though MGB-2 (1200hrs) has a much larger volume of data than CV11 (89hrs). This is also related to the domains covered by CV11 being more open than MGB-2. To further investigate the domain's effect, we juxtaposed the outcomes of \textit{whisper-lg-v2} from scenario-1 with those of \textit{whisper-msa} and \textit{whisper-mix} from scenario-2. It was observed that \textit{whisper-lg-v2} outperformed both models across all dialects, despite being the foundational model for the latter two. However, in the case of \textit{whisper-egyptian} and \textit{whisper-morrocan}, each surpassed \textit{whisper-lg-v2} within their respective dialects as well as in Algerian with the Morrocan model. These findings highlight the significance of incorporating models that are both open-domain and dialect-specific. Moreover, they highlight a clear gap between the current multilingual and SOTA Arabic models on one hand, and actual world dialects on the other. We hope that~\ourdataset~contributes to bridging this gap.

To further explore the effectiveness of \ourdataset, we fine-tune Whisper-v3 using combined training splits from each dialect (\textit{Whisper-Casablanca}) and conducted an evaluation on the Algerian dialect as a case study. We compare this model to \textit{Whisper-lg-v3} as the baseline, \textit{Whisper-mixed}, which was pre-trained on the largest dataset, and \textit{Whisper-Moroccan}, the top-performing model for the Algerian dialect. The results displayed in Table~\ref{tab-alg-result} demonstrate a notable performance improvement over previous models. In comparison with \textit{Whisper-Moroccan}, \textit{Whisper-Casablanca} shows a $14.06$ point reduction in WER before preprocessing and a $16.55$ point reduction after preprocessing. 

\begin{table}[h!]
\resizebox{0.5\textwidth}{!}{%
\begin{tabular}{p{4.6cm}ll}
\hline
\textbf{Model} & \textbf{- Pre-proc}  & \textbf{+ Pre-proc} \\ \hline
Whisper-lg-v3 & 83.49 / 40.47 & 84.14 / 39.99 \\
Whisper-mixed               & 129.63 / 79.63	& 129.77 / 79.68                   \\ 
Whisper-Morrocan  & 74.39 / 29.50 & 74.40 / 29.42                   \\
\textit{Whisper-Casablanca}               & \textbf{60.33 / 26.92}	&  \textbf{57.85 / 25.38} 
                \\ \hline
\end{tabular}
}
\caption{Results for evaluating different Whisper models on the Algerian Test set. Results are reported in WER and CER (/
separated). \textbf{pre-proc:} preprocessing (+ with, - without).
}
\label{tab-alg-result}
\end{table}

\subsection{Evaluation on Code-Switched Data Only}\label{subsec:cse}
For \textit{\textbf{code-switching evaluation}}, we specifically focused on \textit{whisper-large-v3}, selected for its overall superior performance compared to other models, as aforementioned (See Table~\ref{tab-sc-1}). We conducted evaluations first on the original segments containing code-switching with Latin characters, and subsequently on their transliterated counterparts. Due to the relatively small number of code-switching segments, we consolidated all instances into one collective set for this focused evaluation. In the experiments, we evaluated Whisper's performance with inputs featuring either code-switching \textit{(CS-)} or transliteration \textit{(Transliterated-)}, under three distinct decoding scenarios: (1) decoding without specifying the language \textit{(-Auto)}, (2) decoding with English identified as the language \textit{(-EN)}, and (3) decoding with Arabic recognized as the language \textit{(-AR)}. 
%
%
\begin{table}[h!]
\resizebox{0.5\textwidth}{!}{%

\begin{tabular}{p{6.3cm}r}
\hline
\textbf{Condition-predefined} & \textbf{WER / CER} \\ \hline
CS-Auto & 90.89 / 56.72 \\
Transliterated-Auto               & 90.39 / 52.79                   \\ 
CS-EN  & 131.54 / 108.07                   \\
Transliterated-EN               & 133.48 / 115.56                   \\ 
CS-AR   & 103.57 / 67.58 \\
Transliterated-AR              & 100.47 / 58.35                    \\ \hline
\end{tabular}
}
\caption{Evaluation results for \textit{whisper-lg-v3} on the segments with code-switching (Latin characters [CS]), and on the transliterated versions (Transliterated). Prefix \textbf{CS}: reference written with code-switching. Prefix \textbf{Transliterated}: reference written with Arabic letters. Postfix \textbf{Auto}: results without
identifying the decoding language. Postfix \textbf{EN}: results with identifying the decoding language as English. Postfix \textbf{AR}: results with identifying the decoding
language as Arabic.
}
\label{tab-cs-result}
\end{table}
As reported in Table~\ref{tab-cs-result}, the WER/CER scores are high in all settings, however identifying the target language makes the prediction worse. For a deeper comprehension of these findings, Table~\ref{tab-cs-analysis1} and Table~\ref{tab-cs-analysis2} detail the outputs for each condition, specifically for inputs involving code-switching and transliteration, respectively. With code-switched inputs, Table~\ref{tab-cs-analysis1}, Whisper failed to produce any code-switched words in all scenarios. Notably, even when the decoding language was set to English, Whisper performed a translation task even when specifying the task as \textit{"transcription"}. For the Auto and Arabic settings, Whisper outputted only transliterations. This issue is also observable with the transliterated inputs, see Table~\ref{tab-cs-analysis2}. This highlights a limitation in Whisper's capacity to transcribe data containing code-switching.

\subsection{Evaluation on Other Tasks}\label{subsec:eval-speech}
In addition to the main ASR evaluations, we also performed a zero-shot benchmark on two additional tasks: Arabic dialect identification (ADI) and gender recognition. For ADI, we use the best-performing HuBERT-based model from \cite{sullivan2023robustness} and perform a zero-shot evaluation on Casablanca's eight dialects. The results in Table \ref{tab-adi-gen-result} reflect similar challenges observed in their study, where the model underperformed on the "YouTube Dramas" domain. In addition to providing dialect labels, Casablanca also includes gender information, as mentioned in Section \ref{subsec-tasks}. This allows for an evaluation of the gender recognition task. Therefore, we fine-tuned XLS-R \cite{babu2021xls} on Librispeech-clean-100 \cite{panayotov2015librispeech}, as an out-of-domain dataset\footnote{Read-out books also trained on different language (i.e., English).}, and subsequently evaluated its performance on our dataset. 



\begin{table}[h!]
\resizebox{0.5\textwidth}{!}{%

\begin{tabular}{p{2.1cm}llll}
\hline
\textbf{Task} & \textbf{Accuracy}  & \textbf{Precision} & \textbf{Recall} & \textbf{F1 Score} \\ \hline
ADI & 36.44 & 54.68 & 36.44 & 39.24 \\
Gender Rec.& 83.56 	& 89.23 & 83.56 & 84.32 \\ 
\hline
\end{tabular}
}
\caption{Zero-shot results of ADI and gender recognition tasks on~\ourdataset.
}
\label{tab-adi-gen-result}
\end{table}


%% file: Sections/Conclusion.tex
\section{Conclusion}
\label{sec:conclusion}
In this paper, we introduced \ourdataset, the largest supervised dataset for Arabic dialects, featuring a diverse representation across eight dialects.~\ourdataset~includes underrepresented dialects such as Emirati, Yemeni, and Mauritanian. Encompassing 48 hours of data, the dataset also involves detailed annotations on transcriptions, speaker gender, and code-switching. Initial experiments with SoTA models demonstrate the ~\ourdataset's utility for enhancing Arabic speech processing, especially in ASR, gender identification, and dialect identification. A subset of \ourdataset\ is publicly available, aiming to support further research and innovation in both speech processing as well as linguistic research targeting dialects.


%% file: Sections/Limitations.tex
\section{Limitations}

While we believe~\ourdataset\ will have a significant impact on a wide range of tasks in Arabic speech, it is important to acknowledge some limitations. Although~\ourdataset\ includes eight dialects, substantially more than previous datasets, the Arabic language comprises several other dialects that we do not cover. In addition to dialects, there is also diversity within each dialect.\footnote{If we go by country level, we can talk about 22 dialects. However,~\citet{mageed2020toward} also introduce the concept of micro-dialects to describe sub-country variation.} Therefore, we hope to expand the dataset to encompass a broader range of dialects in the future. Furthermore, as Figure \ref{fig:map} illustrates, for all dialects, the majority of speakers in~\ourdataset\ are male (over 60\%, except for Morocco), potentially introducing gender biases. We recommend caution when working with gender-sensitive tasks. Finally, we provide only a YouTube URL for the source videos instead of the videos themselves due to copyright considerations. This could lead to availability issues if the videos are removed by their authors.


%% file: Sections/Ethical_Statement.tex
\section{Ethical Considerations}
In developing~\ourdataset, we adhere to ethical principles to ensure responsible and respectful use of data. Our dataset, sourced from publicly available TV series episodes on YouTube, is curated with careful consideration for privacy, omitting any personal identifiable information beyond what is publicly accessible. We try our best to ensure diverse representation in terms of gender and dialects to mitigate biases and promote inclusivity in ASR systems. All annotations and evaluations were conducted with linguistic and cultural sensitivity. While aiming to share the dataset to advance research, we implement access policies that require responsible use and proper citation. Our commitment to ethical standards is ongoing, and we welcome community feedback to continuously improve our practices.

%% file: ack.tex
\section*{Acknowledgments}\label{sec:acknow}
We acknowledge support from Canada Research Chairs (CRC), the Natural Sciences and Engineering Research Council of Canada (NSERC; RGPIN-2018-04267), the Social Sciences and Humanities Research Council of Canada (SSHRC; 895-2020-1004; 895-2021-1008), Canadian Foundation for Innovation (CFI; 37771), Digital Research Alliance of Canada\footnote{\href{https://alliancecan.ca}{https://alliancecan.ca}}, and UBC Advanced Research Computing-Sockeye\footnote{\href{https://arc.ubc.ca/ubc-arc-sockeye}{https://arc.ubc.ca/ubc-arc-sockeye}}.

%% file: Sections/APPENDIXES/Appendix.tex
\input{Sections/APPENDIXES/Related_work_Speech}
\input{Sections/APPENDIXES/annotation_tool}
\input{Sections/APPENDIXES/dialects_variations}

\input{Sections/APPENDIXES/special_cases}
\input{Sections/APPENDIXES/Preprocessing}

\input{Sections/APPENDIXES/CS}

%% file: Sections/APPENDIXES/Related_work_Speech.tex
\subsection{Arabic ASR}
\label{app:ar-speech}
Historically, the Hidden Markov Model (HMM) combined with Gaussian Mixture Models (GMM) has been the dominant approach for achieving top results in large vocabulary continuous speech recognition (LVCSR). The first HMM-DNN hybrid for LVCSR was introduced by \citet{dahl2011context}, outperforming traditional HMM-GMM systems. In the MGB2 challenge, \citet{khurana2016qcri} utilized a combination of TDNN, LSTM, and BLSTM models, achieving a notable word error rate (WER) of 14.2\%. End-to-end (E2E) models, mapping speech directly to text, gained popularity, simplifying ASR pipelines. \citet{ahmed2019end} introduced an E2E ASR model for Arabic, leveraging BRNNs with CTC for alignment. The introduction of an E2E transformer model addresses the morphological complexity and
dialectal variations inherent in Arabic using
self-attention mechanism and sub-word tokenization. \citet{hussein2022arabic} advanced Arabic ASR by employing a transformer-based encoder-decoder with a TDNN-LSTM language model, using Mel filter banks for acoustic features and training on MGB3 and MGB5 corpora, achieving leading performance with WERs of 27.5\% for MGB3 and 33.8\% for MGB5. In the era of large speech models, Arabic speech is still in its early stages. The XLS-R model~\cite{babu2021xls}, a large-scale model designed for cross-lingual speech representation learning, utilizing the wav2vec 2.0 framework~\cite{baevski2020wav2vec}, was utilized on the Mozilla Common Voice dataset for MSA \cite{zouhair2021automatic, bakheet2021improving}.
The study of~\citet{ardila2019common} benchmarks foundational models on Arabic ASR tasks, focusing on the performance of OpenAI's Whisper~\cite{radford2023robust}, Google's USM~\cite{zhang2023google}, and the  KANARI ASR model. These models were evaluated against a variety of datasets, emphasizing their efficacy across different Arabic dialects and speaking styles. Notably, USM typically surpassed Whisper, while KANARI demonstrated exceptional capability, especially in code-switching contexts between MSA and Egyptian dialect. The performance of Whisper across various Arabic dialects for ASR tasks was explored by \citet{talafha2023n}. This evaluation spanned most publicly available datasets, utilizing n-shot (zero-, few-, full) fine-tuning approaches. The study also assessed Whisper's adaptability to novel scenarios, including dialect-accented MSA and previously unseen dialects. While Whisper demonstrated competitive results with MSA in zero-shot settings, its ability to adjust to different dialects was limited, showing inadequate performance and random output generation when encountering unfamiliar dialects.

%% file: Sections/APPENDIXES/annotation_tool.tex
\subsection{Annotation Tool}
\label{sec-annotation-tool}
We employed \textit{Label-Studio}\footnote{\href{https://labelstud.io/}{https://labelstud.io/}}, a widely supported open-source labeling platform, as our choice for an annotation tool. We centrally hosted it on our servers and provided online access, allowing for remote and adaptable involvement from annotators across various locations. 
Within the tool we used the `\textit{Automatic Speech Recognition using Segments}' template, enabling annotators to select multiple spans from each snippet and write their transcriptions accompanied by additional metadata. We also customized the tool to allow annotators to specify the gender of the speaker for each segment. 
We randomly shuffled the data to guarantee each snippet's independence, effectively reducing potential bias and sequencing effects that could impact annotators' perceptions during the annotation process.



\subsection{Transcribing a segment}
\label{sec-msadiaothers}

\begin{figure}[h!]
  \begin{center}
    \includegraphics[width=\linewidth]{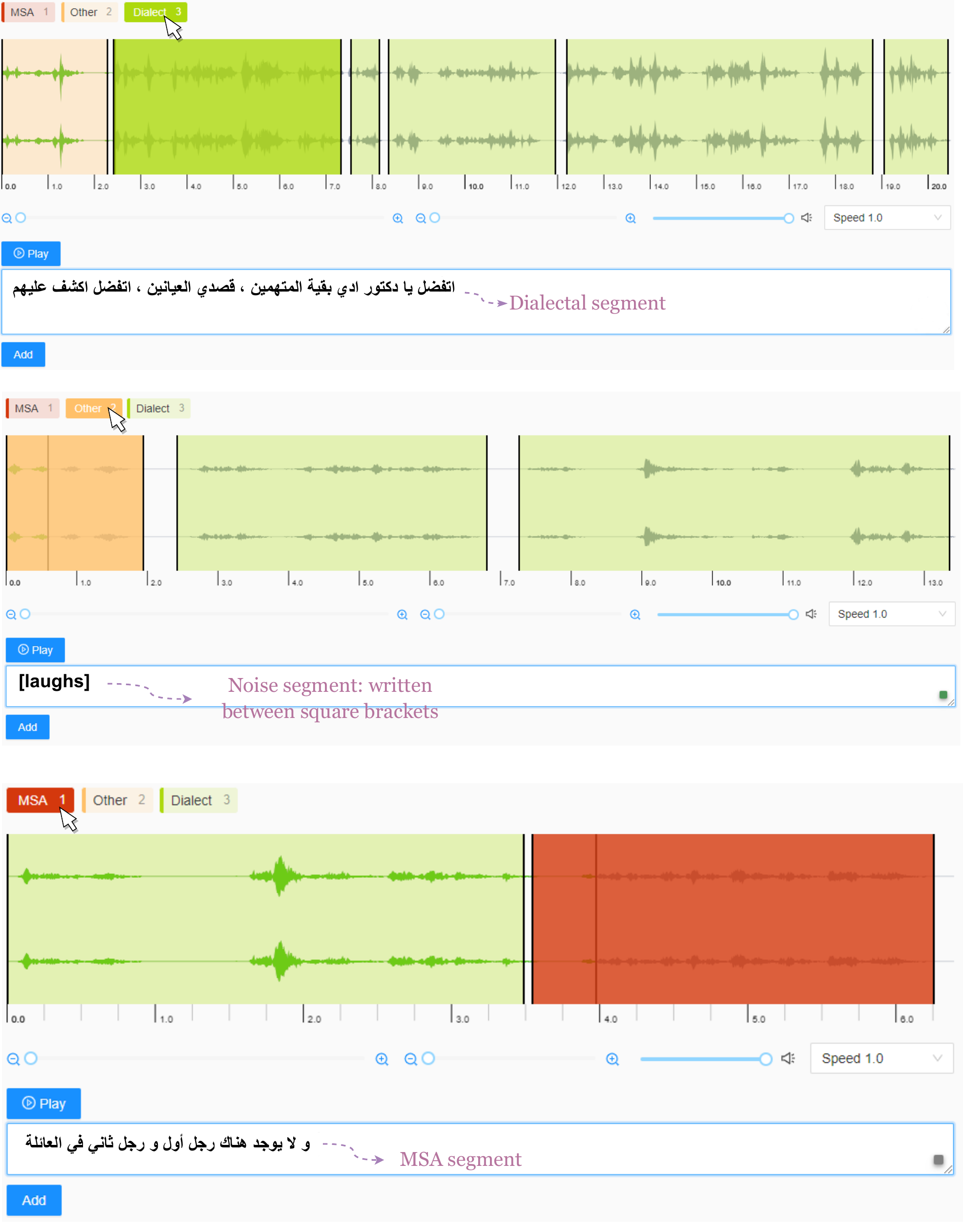}
  \end{center}
\caption{Example of transcribing a segment.
 }
\label{fig:msadiaothers}
\end{figure}

Figure~\ref{fig:msadiaothers} shows the process of transcribing a speech segment from a snippet based on its category (Dialect, MSA, and Other). 

%% file: Sections/APPENDIXES/dialects_variations.tex
\subsection{Inter-dialect diversity}
\label{Inter-dialect}
Table~\ref{tab:variations} demonstrates how the same words can be written differently within the same dialect, showcasing the inter-dialect diversity and the rich nuances that this brings to dialectical expression. 

\begin{table}[h!]
\centering
\resizebox{0.5\textwidth}{!}{%
\def\arraystretch{1.5}
\begin{tabular}{lrrrrl}
\hline
\textbf{Dialect}          & \textbf{Var-1}      & \textbf{Var-1} & \textbf{Var-3} & \textbf{MSA} & \textbf{English}      \\
\hline
\textbf{Algeria} &  \<شوالا>\ & \<واش>  & \<واشنو> & \<ماذا> & What\\

\textbf{Egypt} & \<بردو>\ & \<برضه>  & \<برضو> & \<أيضا> & Also\\
\textbf{Jordan} & 
 \<حكيتله> 
& \<حكيتلو> 
& \<حكيت له> 
& \<قلت له>
& I told him
\\
\textbf{Morocco} & \<عا قوتلو> & \RL{hy gtlyh} & \<غير قلت ليه>  & \<قلت له فقط> & I just told him\\
\textbf{Mauritania} & \<أمبجو> & \<أمبيو> & \<أمبدو> & \<لحاف>& Quilt\\
\textbf{Palestine} & \<هاض> & \<هاد> & \<هاظ\ > & \<هذا> & This\\
\textbf{UAE} & \<قتله> & \<قلتله> & \<قلت له\ > & \<قلت له>
& I told him\\
\textbf{Yemen} & \<ابصرت> & \<ابصرك> & \<ابسرت> & \<أرأيت> & did you see?\\\hline

\end{tabular}
}
\caption{Examples of dialect variation along with their translations in MSA and English. \textbf{Var}: variation.}
\label{tab:variations}
\end{table}

\subsection{Code-switching transcription}
\label{app-cs}

Table~\ref{tab:cs} shows the code-switching transcription process.

\begin{table}[h!]
\centering
\resizebox{0.5\textwidth}{!}{%
\begin{tabular}{ll}
\hline
\textbf{Format}          & \textbf{Transcript}      \\
\hline
Transliterated   &\< أول ما يوصل! أوكي، يلا باي> \\
Untransliterated &

bye
\<يلا >
\<،>
okay 
\<أول ما يوصل!>
 \\
 MSA & 
\< في حين وصوله! حسنًا، مع السلامه>
 \\
English & As soon as he arrives! Okay, bye\\\hline

\end{tabular}
}
\caption{Examples of code-switching in transcription.}
\label{tab:cs}
\end{table}

Table \ref{tab-csstat} shows examples of code-switching segments for each dialect, along with their transliterated versions. Code-switched terms are provided in teal color.

\begin{table}[h!]
\resizebox{0.5\textwidth}{!}{%
\def\arraystretch{1.5}

\begin{tabular}{lr}
\hline
\textbf{Dialect} &
  
  \textbf{Example} 
\\
 \hline
\textbf{Algeria} &
  
\begin{tabular}[c]{@{}r@{}}

\<مخدومة> \textcolor{teal}{l'affaire} \<ماجاتش> \<أوكان> \<لباك> \<يجيب> \<رايح> \<لقمان> \<ويلا> \<علابالكم> \<كيفاه>

\\
 \<مخدومة> 
 \textcolor{teal}{\<لافار>}
 \<ماجاتش> \<أوكان> \<لباك> \<يجيب> \<رايح> \<لقمان> \<ويلا> \<علابالكم> \<كيفاه>
\end{tabular} 
\\

\textbf{Egypt} &
  
\begin{tabular}[c]{@{}r@{}}
\<دلوقتي>. \<من> \<دقايق> \<بعد> \<هيبتدي> \textcolor{teal}{program} \<ال> \<وقتك>. \<في> \<جيت> \<إنت> \<ده> \<أقعد> \<أقعد،>
 
\\
\<دلوقتي>. \<من> \<دقايق> \<بعد> \<هيبتدي>

\textcolor{teal}{\<البروجرام>}

\<وقتك>. \<في> \<جيت> \<إنت> \<ده> \<أقعد> \<أقعد،>
\end{tabular} 
\\

\textbf{Jordan} &
  
\begin{tabular}[c]{@{}r@{}}
\textcolor{teal}{professional} \<يعني> \textcolor{teal}{international} \<الهندي> \<المطرب> \<هذا> \<انو>

\\
\textcolor{teal}{\<بروفيشنال>}
\<يعني> 
\textcolor{teal}{\<انترناشونال>}

\<الهندي> \<المطرب> \<هذا> \<انو>
\end{tabular} 
\\

\textbf{Mauritania} &
  \begin{tabular}[c]{@{}r@{}}
  \textcolor{teal}{Quinze}
 \< بيني و بينك>

\\
\textcolor{teal}{
\<كوينز>
}
 \< بيني و بينك>

\end{tabular}

\\

\textbf{Morocco} &
  
\begin{tabular}[c]{@{}r@{}}
\<دالعرس؟> \textcolor{teal}{préparation} \<ف> \<وصلتي> \<فين>

\\
\<دالعرس؟> 
\textcolor{teal}{\<فبريباراسيون>}
\<وصلتي> \<فين>
\end{tabular} 
\\

\textbf{Palestine} &
  
\begin{tabular}[c]{@{}r@{}}
\<تتجمع> \<الناس> \<ما> \<عبال> \<ثمانية> \textcolor{teal}{maximum} \<سبعة> \textcolor{teal}{maybe} \<يعني>
 
\\
\<تتجمع> \<الناس> \<ما> \<عبال> \<ثمانية>
\textcolor{teal}{\<ماكسيموم>}
\<سبعة>
\textcolor{teal}{\<ميبي>}
\<يعني>
\end{tabular} 
\\

\textbf{UAE} &
  
\begin{tabular}[c]{@{}r@{}}
\textcolor{teal}{fast}
\textcolor{teal}{food} 
\<ال>
\<زمن> \<في> \<عايشين> \<إحنا> \<ألحين> \<أول،> \<مال> \<الحركات> \<هاي>
 
\\
\textcolor{teal}{\<فود> \<الفاست>}

\<زمن> \<في> \<عايشين> \<إحنا> \<ألحين> \<أول،> \<مال> \<الحركات> \<هاي>
\end{tabular} 
\\

\textbf{Yemen} &
  --- 
\\
 \hline

\end{tabular}
}
\caption{Examples of code-switching segments per dialect along with the transliterated version. Code-switched terms are provided in teal color.}
\label{tab-csstat}
\end{table}

%% file: Sections/APPENDIXES/special_cases.tex
\subsection{Special cases}\label{apx:special-cases}
The special cases document served both as a collaborative tool for discussing and standardizing unique dialectal scenarios and as a repository for documenting dialect-specific variations and complex linguistic situations encountered during transcription. Table \ref{tab:specialcases} illustrates some examples.

\begin{table*}[t!]
\resizebox{\textwidth}{!}{%
\def\arraystretch{1.5}
\begin{tabular}{p{3.3cm}p{20cm}}
\hline
\textbf{Dialect} &
  \textbf{Description} \\
  \hline
Egyptian &
  Some speakers tend to use "\<ع>" in the beginning of the words instead of “\<ه>”, so we agreed on writing it as "\<ه>". Others use the letter "\<ح>" as in "\<حقولك>" instead of "\<هقولك>". We suggested writing it the way we hear.

Some segments in the Egyptian dialect include urban upper Egyptian other than the Cairene one, so I wrote it as I heard. For example, a word like "\<أقولك>" in Cairene would be "\<أجولك>" in Upper Egyptian. \\


Jordanian &
The word "\<هسا>" is sometimes pronounced as "\<هسع>", so I transcribe it based on the last letter; if "\<ع>" is clear, I write "\<هسع>" otherwise, I write "\<هسا>".

The word "Tomorrow" has two forms: \<بكرا> and \<بكره>. I decided to write \<بكرا> to be distinguished from \<بكره> which also means "I hate". \\

UAE &

In many pronunciations, some Emaratis (depending on the region and tribe they belong to) put emphasis on some letters in a word. The word "\<علي>" which means on top of me, can also be pronounced with an emphasis on the letter "\<ي>". Another instance is where the letter "\<ه>"  is added at the end of the word "\<عليه>".

Emiratis use the word "\<عيل>" mainly meaning "\<إذا ماذا ؟>"  or what else? However, the word has a less frequent use that means to be the cause of an issue "\<عيل عليه>" or "\<عال عليه>", but with a slightly different pronunciation.
\\
\hline
\end{tabular}
}
\\
\caption{Illustrations of special cases unique to each dialect.}
\label{tab:specialcases}

\end{table*}

%% file: Sections/APPENDIXES/Preprocessing.tex
\subsection{Preprocssing \& settings}
\label{apx:preproc}
For all experiments, we utilize \textit{transformers}\footnote{\href{https://huggingface.co/docs/transformers/index}{https://huggingface.co/docs/transformers/index}} and \textit{datasets}\footnote{\href{https://huggingface.co/docs/datasets/index}{https://huggingface.co/docs/datasets/index}} libraries to load the models and datasets, respectively. We resample all audio segments to a 16kHz rate and perform the text preprocessing steps. We use a single node with A100-SXM4-40GB GPU for all evaluations. 
During the evaluation, we determine the WER  and CER using the original reference and predicted transcriptions. Additionally, we apply text preprocessing to both the reference texts and predictions, adhering to the procedures outlined in~\newcite{talafha2023n}. Specifically, we: (a) retain only the \% and @ symbols, removing other punctuation; (b) eliminate diacritics, Hamzas, and Maddas; and (c) convert Eastern Arabic numerals to Western Arabic numerals (for instance, \<٩٢> becomes 29). We keep all Latin characters as we have code-switching in~\ourdataset.

%% file: Sections/APPENDIXES/CS.tex
\subsection{Code-switching analysis}
To further understand code-switching evaluation, Tables \ref{tab-cs-analysis1} and \ref{tab-cs-analysis2} provide detailed outputs for each condition (see Section \ref{subsec:cse}), focusing specifically on inputs involving code-switching and transliteration, respectively. We use whisper-lg-v3 for all conditions.

\begin{table}[t!]
\resizebox{0.5\textwidth}{!}{%
\def\arraystretch{1.5}
\begin{tabular}{lr}
\hline
\multicolumn{2}{c}{\textbf{Code-switching   input}}                                                   \\ \hline
\multirow{4}{*}{\textbf{CS\_reference}} & \<تتجمع> \<الناس> \<ما> \<عبال> \<ثمانية> maximum \<سبعة> maybe \<يعني>                              \\
                                      & \<ثمانية> Maximum \<و> \<السبعة> \<بين>                                  \\
                                               & signature. la \<ف> \<الحق> \<عندكش> \<ما> \<بالحق> \<فالشركة>. \<النص> \<عندك> \<بلي> \<عارفة>               \\
                                      & \<كصديق> \<إلا> \<لك> \<نظرت> \<ما> \<عمري> \<أنا> \<سامر،> sorry                    \\[0.6cm]
\multirow{4}{*}{\textbf{CS -   Auto}} & 8 \<ماكسيموم> 7 \<ميبو> \<يعني>                                  \\
                                      & \<ثمان> \<وماكسوم> \<السبعة> \<بين>                                  \\
                                      & \<سينياتروخ> \<لا> \<في> \<حق> \<عندكش> \<ما> \<بالحق> \<الشركة> \<نصف> \<عندك> \<بلي> \<عارفة>\\
                                      & \<كصديق> \<إلا> \<لك> \<نظرت> \<ما> \<عمري> \<أنا> \<سامر> \<ساري>                     \\[0.6cm]
\multirow{4}{*}{\textbf{CS - EN}}     & Maybe 7, maximum 8                                            \\
                                      & between 7 and maximum 8                                       \\
                                               & I know you have half the company. You don't   have the right to have a seniority. \\
                                      & Sorry, Samer. I've never seen you except as   a friend.       \\[0.6cm]
\multirow{4}{*}{\textbf{CS - AR}}     & 8 \<ماكسيموم> 7 \<ميبو> \<يعني>                                     \\
                                      & \<ثمان> \<وماكسوم> \<السبعة> \<بين>                                  \\
                                      & \<سينياتروخ> \<لا> \<في> \<حق> \<عندكش> \<ما> \<بالحق> \<الشركة> \<نصف> \<عندك> \<بلي> \<عارفة> \\
                                      & \<كصديق> \<إلا> \<لك> \<نظرت> \<ما> \<عمري> \<أنا> \<سامر> \<ساري>                    \\ \hline
\end{tabular}
}
\caption{
Results of \textit{whisper-lg-v3} on input having code-switching (Latin letters). \textbf{CS\_reference}: reference transcriptions witch code-switching. \textbf{CS - Auto}: output from \textit{whisper-lg-v3} without identifying the decoding language. \textbf{CS - EN}: output from \textit{whisper-lg-v3} with identifying the decoding language as English. \textbf{CS - AR}: output from \textit{whisper-lg-v3} with identifying the decoding language as Arabic.
}
\label{tab-cs-analysis1}
\end{table}


\begin{table}[h]
\resizebox{0.5\textwidth}{!}{%
\def\arraystretch{1.5}
\begin{tabular}{lr}
\hline
\multicolumn{2}{c}{\textbf{Transliterated   input}}                                                                \\ \hline
\multirow{4}{*}{\textbf{\begin{tabular}[c]{@{}l@{}}Transliterated\\ reference\end{tabular}}} &
  \<تتجمع> \<الناس> \<ما> \<عبال> \<ثمانية> \<ماكسيموم> \<سبعة> \<ميبي> \<يعني> \\
                                                & \<ثمانية> \<ماكسيموم> \<و> \<السبعة> \<بين>                                  \\
                                                & \<سينياتور>. \<فلا> \<الحق> \<عندكش> \<ما> \<بالحق> \<فالشركة>. \<النص> \<عندك> \<بلي> \<عارفة> \\
                                                & \<كصديق> \<إلا> \<لك> \<نظرت> \<ما> \<عمري> \<أنا> \<سامر،> \<سوري>                      \\[0.6cm]
\multirow{4}{*}{\textbf{Transliterated - Auto}} & 8 \<ماكسيموم> 7 \<ميبو> \<يعني>                                        \\
                                                & \<ثمان> \<وماكسوم> \<السبعة> \<بين>                                         \\
                                                & \<سينياتروخ> \<لا> \<في> \<حق> \<عندكش> \<ما> \<بالحق> \<الشركة> \<نصف> \<عندك> \<بلي> \<عارفة>    \\
                                                & \<كصديق> \<إلا> \<لك> \<نظرت> \<ما> \<عمري> \<أنا> \<سامر> \<ساري>                          \\[0.6cm]
\multirow{4}{*}{\textbf{Transliterated - EN}}   & Maybe 7, maximum 8                                               \\
                                                & between 7 and maximum 8                                          \\
 &
  I know you have half the company. You don't   have the right to have a seniority. \\
                                                & Sorry, Samer. I've never seen you except as   a friend.          \\[0.6cm]
\multirow{4}{*}{\textbf{Transliterated - AR}}   & 8 \<ماكسيموم> 7 \<ميبو> \<يعني>                                       \\
                                                & \<ثمان> \<وماكسوم> \<السبعة> \<بين>                                         \\
                                                & \<سينياتروخ> \<لا> \<في> \<حق> \<عندكش> \<ما> \<بالحق> \<الشركة> \<نصف> \<عندك> \<بلي> \<عارفة>    \\
                                                & \<كصديق> \<إلا> \<لك> \<نظرت> \<ما> \<عمري> \<أنا> \<سامر> \<ساري>                          \\ \hline
\end{tabular}
}
\caption{
Results of \textit{whisper-lg-v3} on input having transliterated words (Arabic letters). \textbf{Transliterated reference}: reference transcriptions with transliterated words. \textbf{Transliterated - Auto}: output from \textit{whisper-lg-v3} without identifying the decoding language. \textbf{Transliterated - EN}: output from \textit{whisper-lg-v3} with identifying the decoding language as English. \textbf{Transliterated - AR}: output from \textit{whisper-lg-v3} with identifying the decoding language as Arabic.
}
\label{tab-cs-analysis2}
\end{table}

\subsection{Error Analysis of High Error Rates}
In response to the observed high error rates, particularly those exceeding 100 in our evaluations of the \textit{Whisper-mixed} model, we perform error analysis to study the challenges contributing to these errors. This analysis is particularly focused on the Algerian dialect results, where we identify several cases (See Table 14):

\begin{itemize}
\item \textbf{Case 1: }Incorrect Language Base. The model frequently attempted to transcribe dialect-specific phrases by predicting phonetically similar words in MSA, despite their absence in the actual dialogue.

\item \textbf{Case 2:} Inaccurate Translation Over Transcription. There were instances where the model predicted the MSA translation of phrases rather than transcribing the original dialect text. 

\item \textbf{Case 3:} Random Language Interference. The model sometimes generated sentences in completely unrelated languages, despite settings that specify transcription in Arabic. 

\item \textbf{Case 4:} Phonetic Dissimilarity in Short Utterances. Short utterances led to disproportionately high WER when the model generated MSA sentences not phonetically close to the dialect references.

\end{itemize}

\begin{figure}[h]
\centering
\includegraphics[width=0.5\textwidth]{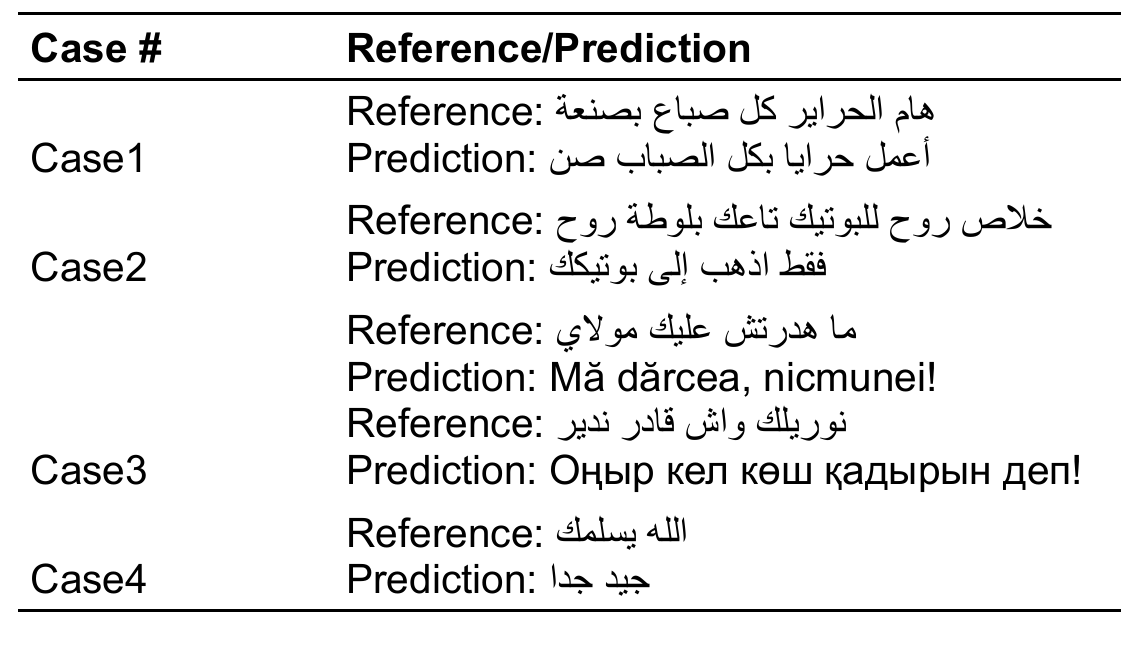}
\caption*{Table 14: Samples from high error rates in the prediction of the Algerian dialect.}
\label{tab:highWER}
\end{figure}